\documentclass{article}

\usepackage[final]{neurips_2024}
\setcitestyle{numbers,square}

\usepackage[utf8]{inputenc}
\usepackage{amsmath,amsfonts}
\usepackage{amssymb}
\usepackage{graphicx}
\usepackage{tikz}
\usepackage{mathtools}
\usepackage{amsthm}
\usepackage{nccmath}
\usepackage{xspace}
\usepackage{xcolor}
\usepackage{makecell}
\usepackage{booktabs}
\usepackage{multirow}
\usepackage{tabularx}
\usepackage{wrapfig}
\usepackage{array}
\definecolor{MyBlue}{rgb}{0.12, 0.12, 0.76}
\usepackage[colorlinks,allcolors=MyBlue]{hyperref}
\usepackage{cleveref}

\usepackage[group-separator={,}]{siunitx}
\usepackage{url}
\usepackage{enumitem}
\usepackage{pdflscape}
\usepackage{verbatim}
\usepackage{arydshln}
\usepackage{multirow}
\usepackage{bigdelim}
\usepackage{float}
\usepackage{pgfplots}
\usepackage[font=small]{caption}
\usepackage{inconsolata}
\usepackage[T1]{fontenc}
\usepackage{subcaption}

\usetikzlibrary{shadows,arrows,decorations,decorations.shapes,backgrounds,shapes,snakes,automata,fit,petri,shapes.multipart,calc,positioning,shapes.geometric,graphs,graphs.standard,plotmarks,math}
\usepackage{tikz-qtree}

\title{Getting By Goal Misgeneralization With a Little Help From a Mentor}

\author{Tu Trinh\thanks{Corresponding author} \\
University of California, Berkeley \\
\texttt{tutrinh@berkeley.edu}
\And Mohamad H. Danesh \\
McGill University \\
\texttt{mohamad.danesh@mail.mcgill.ca}
\And Nguyen X. Khanh\\
University of California, Berkeley \\
\texttt{kxnguyen@berkeley.edu} \\
\And Benjamin Plaut\thanks{Corresponding author} \\
University of California, Berkeley \\
\texttt{plaut@berkeley.edu}
}

\begin{document}

\maketitle

\begin{abstract}

While reinforcement learning (RL) agents often perform well during training, they can struggle with distribution shift in real-world deployments. One particularly severe risk of distribution shift is goal misgeneralization, where the agent learns a proxy goal that coincides with the true goal during training but not during deployment. In this paper, we explore whether allowing an agent to ask for help from a supervisor in unfamiliar situations can mitigate this issue. We focus on agents trained with PPO in the CoinRun environment, a setting known to exhibit goal misgeneralization. We evaluate multiple methods for determining when the agent should request help and find that asking for help consistently improves performance. However, we also find that methods based on the agent's internal state fail to proactively request help, instead waiting until mistakes have already occurred. Further investigation suggests that the agent's internal state does not represent the coin at all, highlighting the importance of learning nuanced representations, the risks of ignoring everything not immediately relevant to reward, and the necessity of developing ask-for-help strategies tailored to the agent's training algorithm.

\end{abstract}

\section{Introduction}
\label{sec:intro}

Reinforcement learning (RL) has successfully enabled artificial intelligence (AI) agents to achieve human-level performance in various applications \cite{levine2016visuomotor, mnih2013atari, schrittwieser2020atari}. However, these applications remain confined to artificial environments \cite{brockman2016openaigym, todorov2012mujoco, tassa2018dmcs} or low-stakes tasks, such as folding bedding \cite{wu2023tidybot}, sorting books \cite{kang2024collabot}, or redirecting customers \cite{ravishankar2024zeroshot}. Deployments in higher-stakes settings such as chemistry labs \cite{darvish2024organa}, construction \cite{feng2014atlas}, or hospital work \cite{agraz2022survey} are much more limited. One reason is that RL agents often behave unpredictably under \textit{distribution shift} \cite{paudel2022distributionshift, fujimoto2024distributionshift}, i.e., when the deployment environment differs significantly from the training environment. Distribution shift is often inevitable due to the real world having dynamics not present in controlled training settings, such interfering humans and agents, materials incompatible with robot appendages, mechanical wear and tear, and more. Some solutions to this include increasing the diversity of training environments or training the agent directly in the real-world, but these can be both costly and risky.

We study a different approach: accept that distribution shift is inevitable but train agents to identify unfamiliar situations and subsequently act cautiously, by refusing to take any action and instead requesting assistance from a supervising human or system. We argue that RL agents must be aware of their own limitations in order to act appropriately and safely in potentially risky real-world scenarios.

\begin{wrapfigure}{r}{0.5\textwidth}
\centering
\begin{subfigure}[t]{0.49\linewidth}
    \centering
    \includegraphics[width=\linewidth]{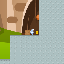}
\end{subfigure}%
\hfill
\begin{subfigure}[t]{0.49\linewidth}
    \centering
    \includegraphics[width=\linewidth]{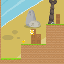}
\end{subfigure}
\caption{Left (training environment): agent successfully gets the coin on the far right. Right (testing environment): agent jumps over the coin in the middle and still heads towards the right.}
\label{fig:coinrun_comparison}
\end{wrapfigure}

These risky real-world scenarios can manifest in many different ways. In this paper, we focus on the phenomenon of goal misgeneralization \cite{langosco2023goalmisgeneralization, shah2022goalmisgeneralization}, in which the agent appears to have successfully learned its task but has actually learned an incorrect goal, one that coincides with the true goal during training but not during deployment. This is particularly nefarious because the agent appears perfectly capable all the way up until its shortcomings are revealed during deployment.
As such, in this paper, we investigate whether asking for help from an expert in anomalous or out-of-distribution test-time situations can mitigate goal misgeneralization. To our knowledge, we are the first to study this issue; details on related work can be found in \Cref{app:related}. We use the CoinRun environment \cite{cobbe2020procgen, langosco2023goalmisgeneralization} for our experiments as it has been shown to induce goal misgeneralization. The agent's goal is to reach a coin, located at the far right during training and at a randomized location during testing, leading the agent to consistently reach the coin during training but completely ignore the coin during testing (see \Cref{fig:coinrun_comparison}).



\section{Experimental Setup}
\label{sec:setup}

\subsection{Training and Testing Environments}
\label{subsec:environments}

Our experiments were conducted using Open\-AI's procgen package \cite{cobbe2020procgen} and a package built on top of it called procgenAISC \cite{langosco2023goalmisgeneralization}. Procgen\-AISC introduces additional difficulty settings for some procgen games that change the procedural generation distribution in significant qualitative ways that lead to goal misgeneralization. We focus on the flagship environment pair used to demonstrate goal misgeneralization: coinrun from \cite{cobbe2020procgen} and coinrun\_aisc from \cite{langosco2023goalmisgeneralization}. In both environment distributions, the agent tries to reach a coin, receiving a reward of 10. All other outcomes such as colliding with a monster or running out of time receive 0 reward.

\subsection{The Cost of Asking For Help}
\label{subsec:cost}

In a real-world application, while asking for help is preferable to harmful behavior, it still incurs a cost: the supervisor's time and effort. Therefore, the agent must balance the cost of harmful behavior with the cost of requesting help. However, since the relative magnitudes of these costs vary greatly across individuals, systems, and applications, we did not incorporate this cost of asking for help into the agent's reward function; rather we only study agent performance as a function of help request frequency. Specifically, we look at the ask-for-help percentage (AFHP), the proportion of timesteps in a run that the agent asks for help.

Each method we test (introduced in Sections \ref{sec:action_methods} and \ref{sec:obs_methods}) uses a different uncertainty metric to determine if it should ask for help. Consequently, each method results in the agent asking for help at different times and a different number of times, yielding a different AFHP for each run. Each run also yields a final reward of either 10 or 0, depending on whether the agent reached the coin. Thus for each method, and for each method's threshold, we compute the realized AFHP and average reward across all runs. Since choosing a specific threshold is arbitrary, we look at each method's performance across all AFHPs. If one method outperforms another for any fixed AFHP, we can conclude that the former is superior. More generally, one could compute the area under the curve of performance on the $y$-axis and AFHP on the $x$-axis, where a larger area indicates that the agent was able to get the coin more often compared to other methods which asked for help at the same rate.


\subsection{Experimentation Pipeline}
\label{subsec:pipeline}

The training and testing pipeline is as follows. We first train an agent using PPO on coinrun. The resulting stochastic policy samples an action from its outputted distribution for the agent to take. We call this the weak agent, as it is expected to suffer from goal misgeneralization as shown in \cite{langosco2023goalmisgeneralization}. To simplify and accelerate the process of giving help, we next train an agent with PPO on coinrun\_aisc. We call this the expert agent as it is a stand-in for a supervisor who knows the optimal actions in the deployment environment. We run the weak agent on coinrun\_aisc and have it ask for help using one of the below methods, upon which the expert agent will provide it with an action to take. Each of these experiments was run on 1000 different environment seeds.


\section{Using Agent's Action Distribution}
\label{sec:action_methods}

We started by using the weak agent’s action distribution as a signal for when to ask for help. We hypothesized that if the agent is in an unfamiliar situation, it would be unsure of what is best to do, which would lead to the action distribution having high uncertainty. We studied five different policies that ask for help based on different measures of uncertainty: (1) max probability: if the probability of the agent's highest-probability action is less than some threshold; (2) max logit: if the highest logit in the agent's action distribution is less than some threshold; (3) sampled probability: if the probability of the agent's sampled (chosen) action is less than some threshold; (4) sampled logit: if the logit corresponding to the agent's sampled (chosen) action is less than some threshold; and (5) entropy: if the entropy of the agent's action distribution is more than some threshold.

\begin{figure}[h]
\centering
\begin{minipage}{0.49\textwidth}
    \centering
    \includegraphics[width=\textwidth]{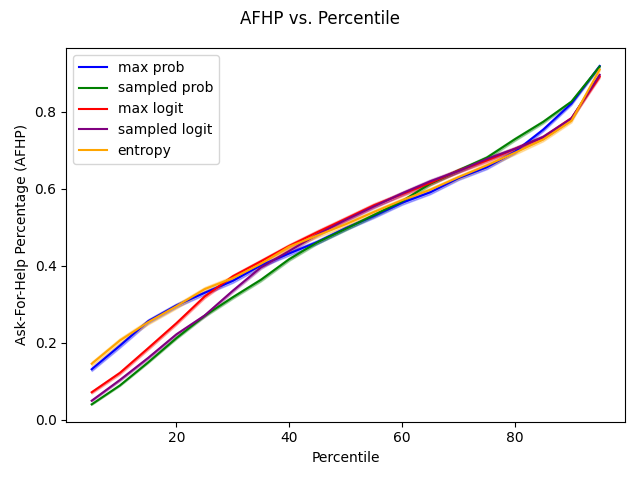}
    \caption{How different percentile thresholds result in different ask-for-help percentages}
    \label{fig:action_method_afhp_by_percentile}
\end{minipage}
\hfill
\begin{minipage}{0.49\textwidth}
    \centering
    \includegraphics[width=\textwidth]{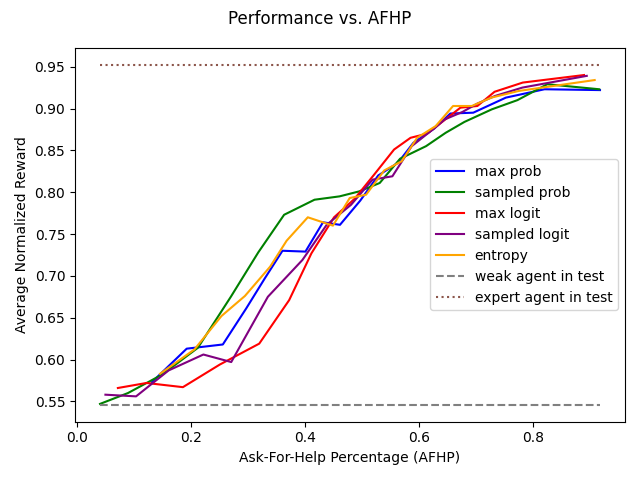}
    \caption{Performance on coinrun\_aisc using action-based methods}
    \label{fig:action_method_performance_by_afhp}
\end{minipage}
\end{figure}

The thresholds for the methods above are created after running the weak agent on coinrun. Using max probability as an example: for every run, every timestep, we recorded the max probability of the agent's action distribution. Then we computed percentiles of these values to use as thresholds. Thus, in practice, the weak agent might ask for help when its current max probability in coinrun\_aisc is less than, for example, the 20th percentile of max probabilities seen in coinrun. As shown in \Cref{fig:action_method_afhp_by_percentile}, a higher percentile corresponds to more frequent help requests—each $p$th percentile in fact results in very close to AFHP = $p$\%.

\Cref{fig:action_method_performance_by_afhp} shows the results of the five action-based methods in coinrun\_aisc. The horizontal lines show the weak agent's performance in coinrun\_aisc (dashed gray) and expert agent's performance in coinrun\_aisc (dotted brown). The action-based ask-for-help methods always outperform the weak agent's no-help performance in the test environment, showing that asking for help based on action distribution uncertainty can mitigate goal misgeneralization, even when the agent asks for help very rarely. There was no significant performance difference between these five methods.

Thorough examination of the agent's runs in coinrun\_aisc revealed that the agent most often asks for help when it reaches the far right wall, the location of the coin during training. One interpretation is that the agent realizes there is no coin to be found and that the level is not ending with high reward as expected. Only then does it repeatedly ask for help. It actually rarely ever asks for help when it encounters the coin in the middle, the actual stark difference with the training environment. We argue that this is not ideal: an effective ask-for-help strategy should be able to recognize anomalies immediately, not wait for the agent to make a mistake before asking, which can be risky, costly, and essentially defeats the purpose of being able to ask for help in the first place.

\section{Using Anomalous Observations}
\label{sec:obs_methods}

This realization motivated us to try observation-based methods instead of action-based methods.
Simply put, if we desire the agent to recognize an anomalous feature of the environment immediately when it appears, then we should work with the environment observations, whether as raw images or latent representations. In order to determine if an observation warranted asking for help, we used the state-of-the-art anomaly detection method Deep-SVDD \cite{ruff2018deepsvdd}. Details on how the Deep-SVDD model was used can be found in \Cref{app:svdd}. We tested two different observation input types to the Deep-SVDD model: (1) raw: look at the raw observation image to determine anomalies; and (2) latent: look at the weak agent's policy's latent representation of the observation (i.e. the environment from the agent's ``point-of-view''), taken from its penultimate layer, to determine anomalies.

We hypothesized that the Deep-SVDD model would give higher anomaly scores to observations with the coin in the middle which never appear during training, thus allowing the agent to ask for help as soon as it recognizes the misplaced coin. Like with the action-based methods above, we use percentiles on the anomaly scores of observations seen in training as thresholds for Deep-SVDD.

\begin{figure}[h]
\centering
\begin{minipage}{0.49\textwidth}
    \centering
    \includegraphics[width=\textwidth]{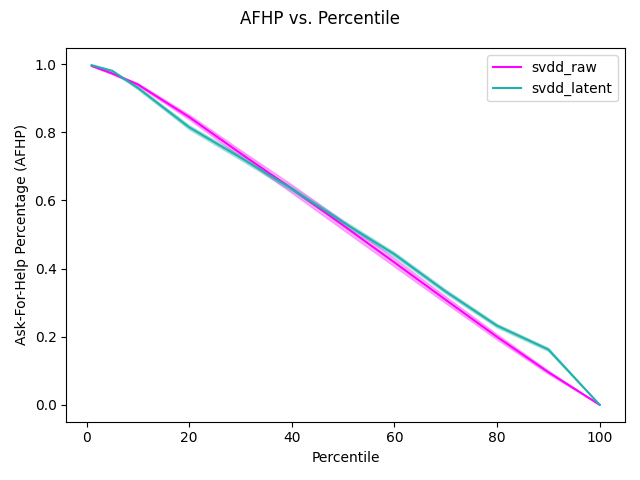}
    \caption{Higher percentile thresholds correspond to being more ``lenient'' with observation abnormality, leading to lower AFHPs}
    \label{fig:obs_method_afhp_by_percentile}
\end{minipage}
\hfill
\begin{minipage}{0.49\textwidth}
    \centering
    \includegraphics[width=\textwidth]{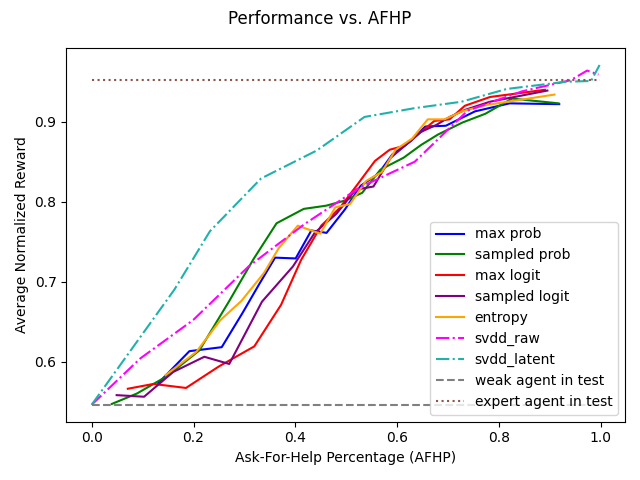}
    \caption{Performance on coinrun\_aisc using observation-based methods, compared to that of action-based methods}
    \label{fig:obs_method_performance_by_afhp}
\end{minipage}
\end{figure}

\begin{wrapfigure}{r}{0.5\textwidth}
    \centering
    \includegraphics[width=0.5\textwidth]{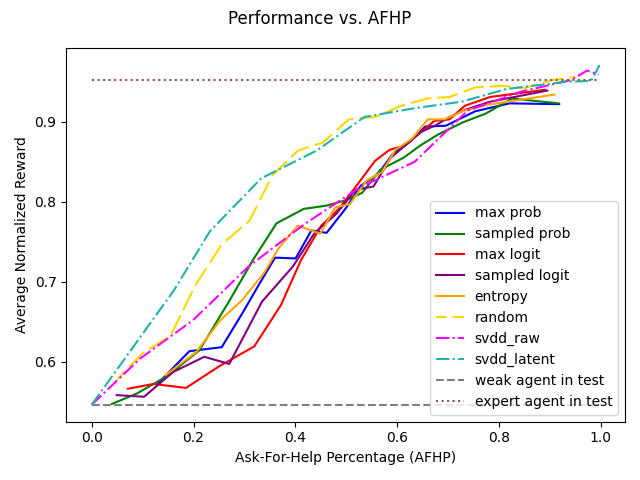}
    \caption{Performance of action-based methods, observation-based methods, and the random baseline on coinrun\_aisc}
    \label{fig:performance_by_afhp_with_random}
\end{wrapfigure}

\Cref{fig:obs_method_performance_by_afhp} shows that the observation-based methods outperform the action-based methods (and trivially the no-help baseline). Using latent observation representations has a clear benefit, but using raw observations only barely outperforms the action-based methods. One possible explanation is that it is difficult to detect different coin locations in a raw image since the coin takes up very few pixels compared to other aspects, such as background pattern, ground colors, monsters, and more.

Unfortunately, we further discovered that neither the action-based nor the observation-based methods outperform a very important baseline: the random baseline, defined by the agent asking for help with some fixed probability at every timestep, independent of the current observation. \Cref{fig:performance_by_afhp_with_random} shows that the random baseline significantly outperforms our action-based methods while performing approximately equally well to our observation-based methods. 

We suspect that the random baseline results in the agent asking for help more in the beginning and middle of the level instead of waiting to reach the right wall at the end or even waiting to approach the coin. This means that the agent has a higher chance of receiving expert guidance early on, \textit{before} it has a chance to make a mistake, leading to a higher frequency of getting the coin as the expert can easily guide the agent forward to the right spot.
The Deep-SVDD-latent method, which has the most similar performance, also results in the agent asking for help throughout the level instead of just at the end, but we did not find any correlation between appearance of the coin in the middle and when this agent asked for help.

\section{Ask-For-Help Skylines and the Importance of Learned Representations}
\label{sec:skyline}

Our discovery that the random baseline induces high-performing, proactive ask-for-help behavior motivated us to ask if there was still much room for improvement for an effective ask-for-help strategy. To answer this, we built \emph{skylines}: ask-for-help methods trained directly on the test environment. The skylines are meta-policies with an action space of size 2 (use the weak agent's action or use the expert agent's action) that try to learn the best cooperation dynamic between the two agents. We emphasize that these skylines are purely theoretical and not viable agent policies, as access to the test distribution is unrealistic for most applications. This is in contrast to our previous methods, which use no knowledge of the test environment.


\begin{wrapfigure}{r}{0.5\textwidth}
    \centering
    \includegraphics[width=0.5\textwidth]{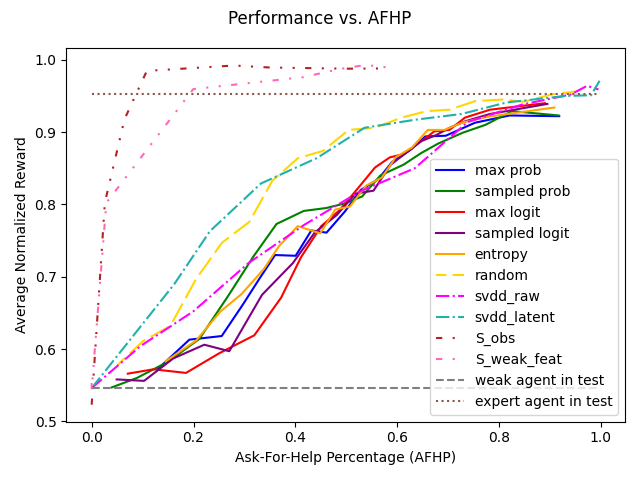}
    \caption{Skyline performances on coinrun\_aisc, compared to previous methods'}
    \label{fig:skyline_results}
\end{wrapfigure}

We tested two different skyline architectures, both trained with PPO: (1) S\_obs: this skyline takes in the raw environment observation image as input to its (meta) policy; and (2) S\_weak\_feat: this skyline takes in the weak agent policy's latent observation representation as input.
\Cref{fig:skyline_results} shows the two different skyline performances on coinrun\_aisc. We can see that they profoundly outperform all other ask-for-help methods, including the random baseline and even the expert agent in its own training environment.
These results answered definitively that there was significant room for improvement before our methods could reach optimal performance. Yet the question still remained: why couldn't the previous methods recognize that having a coin in the middle of the level was an anomaly?

A qualitative analysis revealed key insights into the skylines' behavior. S\_obs asks for help consistently when the coin appears in the middle of the level. 
On the other hand, it appears that S\_weak\_feat's ask-for-help policy does not depend at all on the coin's location! Rather, it tends to asks for help near ledges to determine if the agent should jump (and if so, how high or far out to jump) or if it should continue walking. (\Cref{fig:coinrun_comparison} (right) is an example of the agent jumping over the coin on a ledge if it doesn't ask for help.) Indeed, we found that S\_obs asks for help, on average, 11.4\% of the time when the coin is present and 0\% of the time when it's absent. S\_weak\_feat, meanwhile, only asks for help 6.1\% of the time when the coin is present and 13.4\% of the time when it's absent. We confirmed that a sizable amount of the skyline training data has the coin on a ledge. That means asking for help near a ledge is an excellent ``coin-agnostic'' strategy and one that S\_weak\_feat ultimately decided was the best—if the agent doesn't have any concept of a coin, it would associate ledges with high rewards and seek assistance near them to have the expert lead it to the best landing. 

This analysis suggests an explanation for why methods based on the agent's internal state (action-based methods and Deep-SVDD-latent) are ineffective: these representations do not capture the coin's existence.
Without a comprehensive environment representation, any unsupervised ask-for-help strategy on top of an existing trained agent will be fundamentally limited. With our PPO agents, this suggests that PPO cannot emulate true intelligent agent behavior because it fails to capture elements of the world which are not immediately relevant to the reward. It also suggests that a good ask-for-help strategy \textit{must} be tailored to the specific training algorithm used.

\section{Conclusion}
\label{sec:conclusion}

In this paper we demonstrated that asking for help from a supervisor can mitigate goal misgeneralization in RL agents. However, current methods struggle to proactively detect anomalies, often asking for help only after mistakes have been made. Future work should focus on improving agents' internal environment representations and thus early anomaly detection abilities to fully realize the potential of ask-for-help strategies. Additional areas for future work are discussed in \Cref{app:future_work}.  

\clearpage

\bibliography{bibliography}
\bibliographystyle{plain}

\clearpage

\appendix

\appendix

\section{Social Impact Statement}
\label{app:sis}

Our work focuses on improving the safety and reliability of reinforcement learning agents by enabling them to recognize when they are in unfamiliar situations and act cautiously in response by requesting supervisor assistance. We hope that our work contributes to reducing risks induced by distribution shift and goal misgeneralization, especially in high-stakes real-world applications such as healthcare or assistive robotics.

\section{Deep-SVDD Model}
\label{app:svdd}
The Deep-SVDD model as built by \cite{ruff2018deepsvdd} is trained to form the tightest hypersphere possible around the latent representations of in-distribution datapoints. For our purposes, we trained a Deep-SVDD model on observations collected from running the weak agent in the training environment, coinrun. During testing, the Deep-SVDD model assigns an anomaly score to its input datapoint based on how far its latent representation is from the center of the hypersphere. This score is then compared with a provided threshold to determine if the datapoint is far enough away from the cluster of in-distribution datapoints to be considered anomalous. That means in practice, we feed in the agent's observation from when it's running in coinrun\_aisc into the Deep-SVDD model at each timestep and receive an anomaly score. If this score is above the given threshold, the agent will ask for help. One great feature of the Deep-SVDD model is that it is lightweight and executes rapidly, making it very suitable for assessing every timestep if the agent should ask for help.

\section{Related Work}
\label{app:related}
Several studies have explored RL agents asking for help to improve performance in challenging environments. For instance, Nguyen et al. \cite{nguyen2021learning} introduced a hierarchical ask-for-help framework where agents attempt to learn more about the surroundings of objects they interact with to decompose their tasks into subtasks. Liu et al. \cite{liu2022afk} proposed the asking-for-knowledge (AFK) agent, which generates natural-language questions for agents to discover new knowledge about their environment. Both papers approach the ask-for-help problem by enabling the agent to decrease its uncertainty by increasing its environmental knowledge.

Xie et al. \cite{xie2022proactive} developed a proactive strategy for RL agents to ask for help with states or actions they have predicted to be potentially irreversible. Vandenhof \cite{vandenhof2020asking} treated asking for help as an additional, costly action in the agent's action space and proposed a deep Q-learning algorithm to support it. These methods involve additionally training the RL agent in some way, whether to be adept at detecting unfamiliar situations or balancing exploration, exploitation, and asking for help.

Our approach differs from previous work in two key ways: (1) we aim to study unsupervised methods of asking for help that can potentially be applied on top of a pretrained RL agent, and (2) to the best of our knowledge, we are the first to study asking for help in a goal misgeneralization setting.

\section{Future Work}
\label{app:future_work}
Additional areas where future work can help improve our methods or address some limitations include the following.
\begin{enumerate}
    \item Currently our weak agent only asks for help one timestep at a time. Future work can investigate how the agent can best ask for help for longer periods of time, which can alleviate the cost of switching control back and forth between the supervisor and the agent, a more realistic model of real-world interactions.
    \item Our weak agent currently uses the action provided by the expert to replace its own action when it requests help. However, it would be better if in addition to this, the weak agent is also able to incorporate this knowledge into its own policy, leading to a continuous learning paradigm. This could eventually decrease the necessity of asking for help and supervisor resources as the agent grows more experienced.
\end{enumerate}



\end{document}